\documentclass[11pt,a4paper]{article}
\usepackage[hyperref]{emnlp2020}
\usepackage{times}
\usepackage{latexsym}

\usepackage{graphicx}
\usepackage{microtype}
\usepackage{amsmath}

\aclfinalcopy 

\title{TweetBLM: A Hate Speech Dataset and Analysis of Black\_Lives\_Matter-related Microblogs on Twitter}

\author{Sumit Kumar\thanks{equal contribution} \\
  Birla Institute of Technology\\\
   Mesra, India \\
   \text{sumit.atlancey@gmail.com} \\\And
  
   Raj Ratn Pranesh\footnotemark[1] \\
  Birla Institute of Technology\\
  
   Mesra, India \\
   \text{raj.ratn18@gmail.com} \\

  }

\date{}

\begin{document}
\maketitle
\begin{abstract}

In the past few years, there has been a significant rise in toxic and hateful content on various social media platforms. Recently Black Lives Matter movement came into the picture, causing an avalanche of user-generated responses on the internet. In this paper, we have proposed a Black Lives Matter-related tweet hate speech dataset- TweetBLM. Our dataset comprises 9165 manually annotated tweets that target the Black Lives Matter movement. We annotated the tweets into two classes, i.e., “HATE" and “NON-HATE" based on their content related to racism erupted from the movement for black community. In this work, we also generated useful statistical insights on our dataset and performed a systematic analysis of various machine learning models such as Random Forest, CNN, LSTM, Bi-LSTM, Fasttext, $BERT_{base}$ and $BERT_{large}$ for the classification task on our dataset. Through our work, we aim at contributing to the substantial efforts of the research community for the identification and mitigation of hate speech on the internet. The dataset is publicly available.
\end{abstract}


\section{Introduction}\label{intro}
The rapid development of social media networks and blogging websites has surpassed 3.8 billion active users who use text as a primary mode of digital communication. Researchers have been using Twitter as a pool of evidence for hate speech \cite{garland2020countering} since it has 330 million monthly active users. It also enables one to identify trends within ethnically diverse and disadvantaged audiences. A small percentage of users use discriminatory speech to ridicule and harass particular groups of people based on their gender, ethnicity, sexual identity, or other attributes that have become an obstructive byproduct of social media's development.

The global outbreak of Black Lives Matter protest has resulted in a general disturbance in the personal and social lives of the people. The disruption has resulted in an increased level of anxiety, fear, and an outbreak of sturdy emotions \cite{ahorsu2020fear}. This has led to bitter incidents across the world, for instance, acts of verbal and physical abuse, online harassment, aggression for black communities \cite{ziems2020racism}. The protest started on May 26, 2020, the day after an African-American man, was killed during a police arrest. The Movement peaked on June 6, 2020, and is still undergoing, Reports show that around half a million people turned out in nearly 550 places across the United States.

While it has made efforts to educate about racial justice and counter hate via social media campaigns (e.g. the \#BLM campaign), but their success, effectiveness, and reach remain unclear. Online hate speech has a severe negative impact on the victims, often deteriorating their mental health and causing anxiety \cite{saha2019prevalence}.

Thus, it is critical to study the prevalence and impact of online hate and counter hate speech. In this paper, we presented a new hate speech classification tweet dataset focusing on the Black Life Matter movement. The proposed dataset is consist of tweets having two labels: hate and non-hate.



\textbf{Contribution} The three key contribution in this paper is:
\begin{itemize}
\item{} In this work we have published a novel hate speech detection dataset \ref{data_set} comprising over 9165 manually annotated Black Life Matter tweets in two classes- `Hate' and `Non-Hate'. We presented a statistical analysis on the proposed dataset \ref{sect:analysis}.

\item{} We performed a systematic comparative analysis \ref{result} of various deep learning models for hate detection task on our dataset.
\end{itemize}


\section{Related Work}\label{rel_work}

In the past few years, a lot of research has been done in hate speech detection through natural language processing, which involves analyzing and exploring Hate speech hidden in textual representations. For instance, previous works relied on binary classification such as \cite{kwok2013locate}, \cite{djuric2015hate} and by \cite{nobata2016abusive}.

The foremost challenge in creating a robust hate speech detection system for social media is to have a high-quality dataset. For a machine learning model to understand and generalize on a classification task, it is very important to have an accurately annotated dataset with minimum noise. Most commonly available hate speech classification datasets are comprise binary classes (hate and non-hate) \cite{davidson2017automated}, \cite{gao2017detecting}, \cite{ribeiro2018characterizing}, \cite{qian2019benchmark}. They collected these datasets from large social media platforms such as Gab, Twitter, Facebook, Reedit.

Apart from binary classification datasets, researchers have also worked toward the idea of more fine-grain classification and presented datasets with more specific classes. For example, \cite{rezvan2018quality} focuses on the creation of a quality tweet dataset related to five types of harassment content like sexual, racial, appearance, intellectual, and politics. \cite{waseem2016you}. \cite{jha2017does} proposed the datasets with classes belonging to sexist, racist, benevolent sexism, and hostile sexism.


Multilingual \cite{ousidhoum2019multilingual}, \cite{chung2019conan} and multimodal \cite{gomez2020exploring} hate speech datasets also played a very crucial role in understanding the diversity and extent of hate speech across different languages and modality.

Although previously proposed datasets uniquely contributed to the ongoing hate speech research, much less attention has been given to the detection and analysis of hate speech focusing on the black people's community specifically related to Black Life Matter movement.
Therefore, we created TweetBLM to address this challenge.

\section{Dataset}\label{data_set}
\label{sec:length}
This section will explain the generation process and description of the dataset that we introduced in this paper. We condense the approach for collecting and Pre-processing the user-generated dataset through the tweets to come up with a final dataset. We have summarised the features of the dataset through some examples \ref{table:kys3}, along with the data annotation schemes and guidelines.

\subsection{Data Collection}
\label{sect:pdf}
We crawled Twitter data using the Tweepy\footnote{\href{https://www.tweepy.org/}{https://www.tweepy.org/}} which is a Python library for accessing Twitter Application Programming Interface (API\footnote{\href{https://developer.twitter.com/en/docs/twitter-api/v1/tweets/search/api-reference/get-search-tweets}{https://developer.twitter.com/en/docs/twitter-api/v1/tweets/search/api-reference/get-search-tweets}}), and collected a sample of tweets between 27th May 2020 and 26th July 2020. The sample comprised an average of 1,025,286 million tweets per day. Out of the 7,346,842 tweets, English-written tweets were considered for the dataset, and we selected users having over 150 followers for further steps in order to remove the spam tweets written by bots.

To extract Black Lives Matter movement-related tweets, we build a set of keywords related to the usage of hashtags in the semantic sentence (e.g., \#BLM, \#BlackLivesMatter) in both lowercase and uppercase. The following keywords such as \texttt{\#Atlanta protest}, \texttt{\#BLM}, \texttt{\#ChangeTheSystem}, \texttt{\#JusticeForGeorgeFloyd}, \texttt{\#BlueLivesMatter} were used to collect hate speech specific tweets. Hence, the final collected dataset contains 9165 tweets.

\begin{table*}[]
\begin{tabular}{|c|l|}
\hline
\multicolumn{1}{|l|}{\textbf{Label}} & \multicolumn{1}{c|}{\textbf{Examples}}                                                                                                                                                                                                                                                           \\ \hline
Hate                                    & \begin{tabular}[c]{@{}l@{}}“Between 50-75, \#BlackLivesMatter members surrounded  Officer Joseph Mensah’s girlfriend’s house, \\ assaulted  and physical"\\ \hline



“\#BlackLivesMatter are a bunch of \#terrorists, Trump must win \#BlueLivesMatte \#All-\\ LivesMatter @realDonaldTrump"\\ \hline

“\#BlackLivesMatter Wake Up you are Pawns Being used by the \#Democrats Your Lives\\ mean nothing to them! They are insane"\end{tabular} \\ \hline

Non Hate                                   & \begin{tabular}[c]{@{}l@{}}“\#BlackLivesMatter This should not be a controversial statement. \#BLM"\\
 \hline
“\#BlackLivesMatter Trump supporters are seeding violence. This is also happening in\\ Seattle, Portland, and the areas"\\ 

 
  \hline“\#BlackLivesMatter \#CorneliusFredericks We Demand Justice For Cornelius Fredericks"\\

   \end{tabular}                                                                                                                                                                                                                                           \\ \hline
\end{tabular}
\caption{Example tweets from our collected dataset}
\label{table:kys3}
\end{table*}

\subsection{Data Annotation}
\label{sect:pdf}
We annotated the gathered data based on two categories: Hate and Non-Hate.  The categories were chosen based on the frequency of the occurrence of hate and non-hate text associated with tweets.

We give a general description of each class below. \newline
\textbf{Mention of Hateful text}: This category of tweets containing information that is broadly defined as a form
of expression that “attacks or diminishes, that incites violence or
hate against groups, based on specific characteristics such as physical
appearance, religion, descent, national or ethnic origin, sexual orientation, gender identity, or other,and it can occur with different linguistic styles, even in subtle forms or when humor is used \cite{fortuna2018survey}".

\textbf{Mention of Non-Hateful text}: This category of tweets contains texts that have information that is neutral and doesn't follow the above category or doesn't harm or demeans the emotion and sentiment of a person, group, community, and culture.

Five human annotators of socio-linguistic background and proficiency did annotation of the dataset to detect hate speech related to Black Lives Matter. In this work, we use 5 annotators to annotate the gathered dataset. The annotators were males having age between 20-25, out of which 3 are undergraduate students and 2 are Master's students. To begin with annotations, we collected 11029 tweets from the crawling process. Three annotators annotated each tweet separately. We considered those tweets for our dataset which have 100\% agreement between at least two annotators among the three annotators. And we decided on the final label by the 100\% agreement of the remaining other two annotators. The tweets were deleted if there is no agreement between the remaining other 2 annotators. This gathered data is reliable for performing experiments. Finally, we get 9165 tweets comprising 3084 hate speech and 6081 non-hate speech. The class distribution of the TweetBLM dataset is given in Table \ref{table:ky}.


\begin{table}
\begin{center}
\begin{tabular}{ | c | c| } 
\hline
Class & Number of sentence text \\ 
\hline
 Hate  & 3084   \\ \hline 
Non-Hate  & 6081  \\ 
\hline 
Total & 9165 \\
\hline
\end{tabular}
\end{center}
\caption{Class distribution of TweetBLM datset}
\label{table:ky}
\end{table}
\subsection{Dataset Analysis}\label{sect:analysis}
In this section, we analyzed our collected dataset in order to generate some useful insights. We discovered the top 10 most frequent unigrams, bi-grams, and tri-grams present in the dataset. We also extracted the most frequent hashtags present in the tweets.

\begin{table*}[h]
\centering
\begin{tabular}{|c|c|c|c|c|}
\hline
Model          & \textbf{Accuracy} & \textbf{F1} & \textbf{Precision} & \textbf{Recall} \\ \hline
Random Forest  & 0.773           & 0.775       & 0.785              & 0.767           \\ \hline
LSTM           & 0.762           & 0.767       & 0.775              & 0.759           \\ \hline
BiLSTM         & 0.775           & 0.789       & 0.855              & 0.733           \\ \hline
CNN            & 0.794          & 0.796       & 0.802              & 0.791           \\ \hline
FastText       & 0.827           & 0.829       & 0.833              & 0.825           \\ \hline
$BERT_{base}$  & 0.874           & 0.869       & 0.921              & 0.823           \\ \hline
\textbf{$BERT_{large}$} & \textbf{0.891}           & \textbf{0.889}       & \textbf{0.934}              & \textbf{0.850}           \\ \hline
\end{tabular}
\caption{Performance score of various models.}
  \label{tab:res}
\end{table*}

\paragraph{Data Preprocessing : }\label{sec:p_pro}

Before conducting the analysis and experiments, we preprocessed tweets by firstly converting them to lowercase representation. We also made the tweets free from any unnecessary elements such as username, mentions, links, retweets. We used NLTK\footnote{\href{https://www.nltk.org/}{https://www.nltk.org/}}, a Python module for text processing that removed the English stop words and performed lemmatization of tweets.

\begin{figure}[h]
  \includegraphics[width=1.0\linewidth]{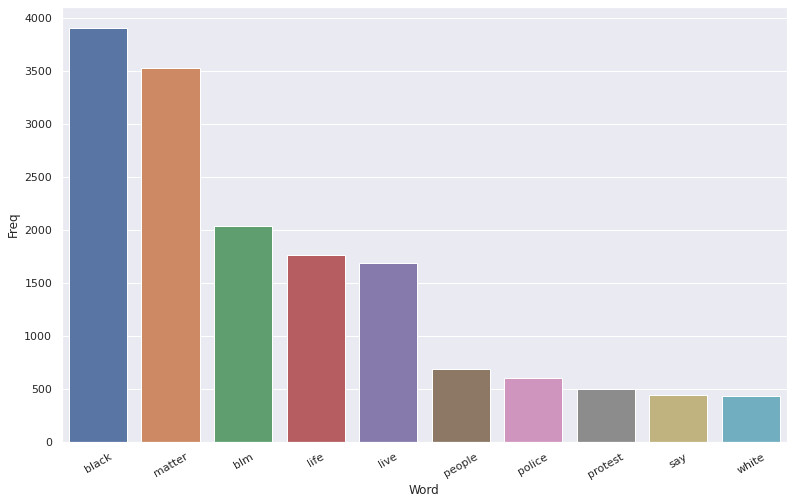}
  \caption{10 most frequent unigram}
  \label{fig:1-gram}
\end{figure}

\begin{figure}[h]
  \includegraphics[width=1.0\linewidth]{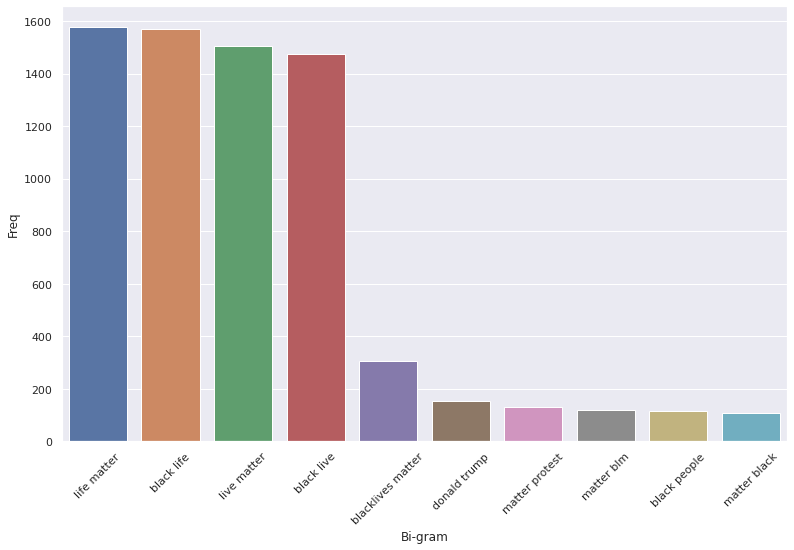}
  \caption{10 most frequent bi-gram}
  \label{fig:2_gram}
\end{figure}

\begin{figure}[h]
  \includegraphics[width=1.0\linewidth]{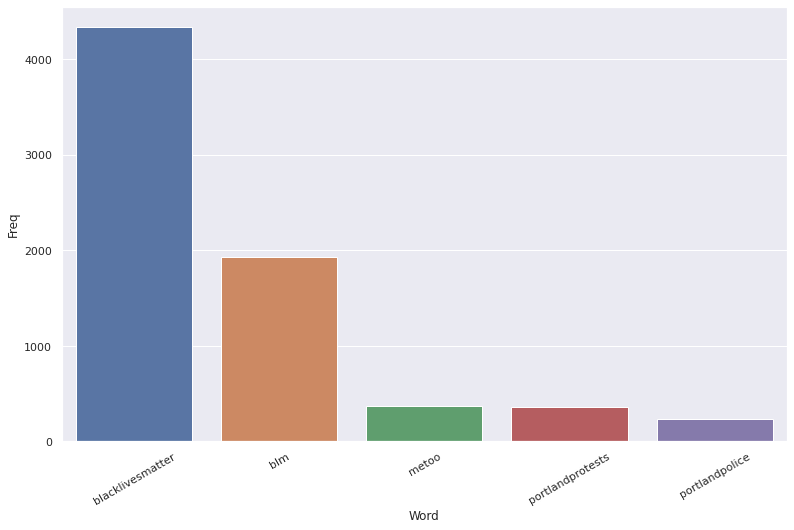}
  \caption{5 most frequent hashtags}
  \label{fig:hash}
\end{figure}

\paragraph{Method : }
Scikit-learn’s CountVectorizer\footnote{\href{https://scikit-learn.org/stable/modules/generated/sklearn.feature_extraction.text.CountVectorizer.html}{https://scikit-learn.org/stable/modules/generated/sklearn\\.feature\_extraction.text.CountVectorizer.html}} module could convert a collection of text documents to a vector of term/token counts. Using CountVectorizer we tokenized the text of the tweet, build a vocabulary of known words, and extracted features from the text of the tweet. As shown in the figure \ref{fig:1-gram} and \ref{fig:2_gram}, we then extracted and visualised the top 10 most frequent unigrams (\textit{black, matter, blm, life, live, people, police, protest, say, white}) and bi-grams(\textit{life matter, black life, live matter, blacklives matter, donald trump, matter protest, matter blm, black people, matter black}) present in our tweet dataset respectively. We also extracted the top 5 most frequently used hashtags(\textit{blacklivesmatter, blm, metoo, portlandprotests and portlandpolice}) present in the tweets(see figure \ref{fig:hash}).

\section{Methodology}\label{method}
In this section, we have sequentially discussed the architecture of various classification models used in our experiment. We used the Random Forest-based classification model. CNN and RNN(LSTM and BiLSTM) based deep learning models. We also used FastText and BERT($BERT_{base}$ and $BERT_{large}$) pretrained encoder for building classification model.



\section{Experiments}\label{experiment}
In this section, we described the experiment setup for various classification models used in the experiment. We divided the tweets into train and test dataset. Out of total 9165 tweets, we used 80\% of the data for training the models and rest 20\% of the data was used for the validation. The hyper-parameters were fine-tuned on the validation dataset. For the \textbf{Random Forest}, we set the \textit{n\_estimators} = 100, \textit{max\_depth} = 5, \textit{max\_features} = 'auto' and rest of the parameters were set to default configuration. For \textbf{CNN} and \textbf{RNN} models, we set the max length(\textit{max\_length}) of input sequence as 120 and used GloVe embedding \cite{pennington2014glove} of size 300 at embedding layer. In \textbf{CNN}, the convolution layer had \textit{filters} = 300, \textit{kernel\_size} = 3, \textit{stride} = 1 with relu activation\cite{nair2010rectified}. For \textbf{RNN} models, we used a \textit{SpatialDropout}(p = 0.2), single LSTM(with dropout = 0.2) and BiLSTM(with dropout = 0.2) layer were for the LSTM and BiLSTM model, respectively. In both CNN and RNN model, a fully connected layer classify the tweet into hate and non-hate class. We used \textit{categorical\_crossentropy} loss with Adam optimizer\cite{kingma2014adam}. For the \textbf{BERT} models- $BERT_{base}$ and $BERT_{large}$, the model had the learning rate of 1e-4 with Adam optimizer\cite{kingma2014adam} and textit{categorical\_crossentropy} as loss function. For the experiment, the model was fine-tuned for 5 epochs. The Random Forest, CNN and RNN models were trained for 25 epochs.

\section{Results and Discussion}\label{result}
In this section, we have summarised the result obtained in our experiment and also discussed the performance of various classification models on our dataset. As seen in the table \ref{tab:res}, the $BERT_{base}$ and $BERT_{large}$ was the best performing models. They could surpass the other models with the highest accuracy score of 87.45\% and 89.13\% respectively. The explanation is that because of the large data used for pretraining of the $BERT_{base}$ and $BERT_{large}$ models and transfer learning increased their contextual understanding and models could generalize better. Following the BERT model, the FastText model could achieve an accuracy of 82.77. With RNN models, BiLSTM performed better than the LSTM model by achieving higher accuracy of 77.58\% which was 1.37\% more than LSTM. The CNN model outperformed both RNN models with an accuracy score of 79.46\%. Random Forest reported an accuracy score of 77.35\%, which was a slight improvement of 1.14\% over the LSTM model.

\section{Conclusions}\label{conclusion}

In this paper, we presented the Black Lives Matter dataset: TweetBLM. We collected 9165 user-generated Black Lives Matter movement-related tweet data which were manually labeled into two classes. We presented statistical insights into our dataset by extracting frequent n-grams and hashtags. Using our dataset, we also performed a comparative analysis of various machine learning models as an attempt to understand how well these models are able to generalize and perform classification on our dataset. 
The experiment results show that one application of our approach might be the identification and filtering of hateful textual contents on social media platforms, but there is still room for improvement. We can do future work such as (i) Expanding our data for annotating more tweets that would be beneficial for the research community, (ii) Providing additional features for fine-grained classification.


\bibliographystyle{acl_natbib}
\bibliography{anthology,emnlp2020}

\begin{thebibliography}{21}
\expandafter\ifx\csname natexlab\endcsname\relax\def\natexlab#1{#1}\fi

\bibitem[{Ahorsu et~al.(2020)Ahorsu, Lin, Imani, Saffari, Griffiths, and
  Pakpour}]{ahorsu2020fear}
Daniel~Kwasi Ahorsu, Chung-Ying Lin, Vida Imani, Mohsen Saffari, Mark~D
  Griffiths, and Amir~H Pakpour. 2020.
\newblock The fear of covid-19 scale: development and initial validation.
\newblock \emph{International journal of mental health and addiction}.

\bibitem[{Chung et~al.(2019)Chung, Kuzmenko, Tekiroglu, and
  Guerini}]{chung2019conan}
Yi-Ling Chung, Elizaveta Kuzmenko, Serra~Sinem Tekiroglu, and Marco Guerini.
  2019.
\newblock Conan--counter narratives through nichesourcing: a multilingual
  dataset of responses to fight online hate speech.
\newblock \emph{arXiv preprint arXiv:1910.03270}.

\bibitem[{Davidson et~al.(2017)Davidson, Warmsley, Macy, and
  Weber}]{davidson2017automated}
Thomas Davidson, Dana Warmsley, Michael Macy, and Ingmar Weber. 2017.
\newblock Automated hate speech detection and the problem of offensive
  language.
\newblock In \emph{Proceedings of the International AAAI Conference on Web and
  Social Media}, volume~11.

\bibitem[{Djuric et~al.(2015)Djuric, Zhou, Morris, Grbovic, Radosavljevic, and
  Bhamidipati}]{djuric2015hate}
Nemanja Djuric, Jing Zhou, Robin Morris, Mihajlo Grbovic, Vladan Radosavljevic,
  and Narayan Bhamidipati. 2015.
\newblock Hate speech detection with comment embeddings.
\newblock In \emph{Proceedings of the 24th international conference on world
  wide web}, pages 29--30.

\bibitem[{Fortuna and Nunes(2018)}]{fortuna2018survey}
Paula Fortuna and S{\'e}rgio Nunes. 2018.
\newblock A survey on automatic detection of hate speech in text.
\newblock \emph{ACM Computing Surveys (CSUR)}, 51(4):1--30.

\bibitem[{Gao and Huang(2017)}]{gao2017detecting}
Lei Gao and Ruihong Huang. 2017.
\newblock Detecting online hate speech using context aware models.
\newblock \emph{arXiv preprint arXiv:1710.07395}.

\bibitem[{Garland et~al.(2020)Garland, Ghazi-Zahedi, Young,
  H{\'e}bert-Dufresne, and Galesic}]{garland2020countering}
Joshua Garland, Keyan Ghazi-Zahedi, Jean-Gabriel Young, Laurent
  H{\'e}bert-Dufresne, and Mirta Galesic. 2020.
\newblock Countering hate on social media: Large scale classification of hate
  and counter speech.
\newblock \emph{arXiv preprint arXiv:2006.01974}.

\bibitem[{Gomez et~al.(2020)Gomez, Gibert, Gomez, and
  Karatzas}]{gomez2020exploring}
Raul Gomez, Jaume Gibert, Lluis Gomez, and Dimosthenis Karatzas. 2020.
\newblock Exploring hate speech detection in multimodal publications.
\newblock In \emph{Proceedings of the IEEE/CVF Winter Conference on
  Applications of Computer Vision}, pages 1470--1478.

\bibitem[{Jha and Mamidi(2017)}]{jha2017does}
Akshita Jha and Radhika Mamidi. 2017.
\newblock When does a compliment become sexist? analysis and classification of
  ambivalent sexism using twitter data.
\newblock In \emph{Proceedings of the second workshop on NLP and computational
  social science}, pages 7--16.

\bibitem[{Kingma and Ba(2014)}]{kingma2014adam}
Diederik~P Kingma and Jimmy Ba. 2014.
\newblock Adam: A method for stochastic optimization.
\newblock \emph{arXiv preprint arXiv:1412.6980}.

\bibitem[{Kwok and Wang(2013)}]{kwok2013locate}
Irene Kwok and Yuzhou Wang. 2013.
\newblock Locate the hate: Detecting tweets against blacks.
\newblock In \emph{Twenty-seventh AAAI conference on artificial intelligence}.

\bibitem[{Nair and Hinton(2010)}]{nair2010rectified}
Vinod Nair and Geoffrey~E Hinton. 2010.
\newblock Rectified linear units improve restricted boltzmann machines.
\newblock In \emph{ICML}.

\bibitem[{Nobata et~al.(2016)Nobata, Tetreault, Thomas, Mehdad, and
  Chang}]{nobata2016abusive}
Chikashi Nobata, Joel Tetreault, Achint Thomas, Yashar Mehdad, and Yi~Chang.
  2016.
\newblock Abusive language detection in online user content.
\newblock In \emph{Proceedings of the 25th international conference on world
  wide web}, pages 145--153.

\bibitem[{Ousidhoum et~al.(2019)Ousidhoum, Lin, Zhang, Song, and
  Yeung}]{ousidhoum2019multilingual}
Nedjma Ousidhoum, Zizheng Lin, Hongming Zhang, Yangqiu Song, and Dit-Yan Yeung.
  2019.
\newblock Multilingual and multi-aspect hate speech analysis.
\newblock \emph{arXiv preprint arXiv:1908.11049}.

\bibitem[{Pennington et~al.(2014)Pennington, Socher, and
  Manning}]{pennington2014glove}
Jeffrey Pennington, Richard Socher, and Christopher~D Manning. 2014.
\newblock Glove: Global vectors for word representation.
\newblock In \emph{Proceedings of the 2014 conference on empirical methods in
  natural language processing (EMNLP)}, pages 1532--1543.

\bibitem[{Qian et~al.(2019)Qian, Bethke, Liu, Belding, and
  Wang}]{qian2019benchmark}
Jing Qian, Anna Bethke, Yinyin Liu, Elizabeth Belding, and William~Yang Wang.
  2019.
\newblock A benchmark dataset for learning to intervene in online hate speech.
\newblock \emph{arXiv preprint arXiv:1909.04251}.

\bibitem[{Rezvan et~al.(2018)Rezvan, Shekarpour, Balasuriya, Thirunarayan,
  Shalin, and Sheth}]{rezvan2018quality}
Mohammadreza Rezvan, Saeedeh Shekarpour, Lakshika Balasuriya, Krishnaprasad
  Thirunarayan, Valerie~L Shalin, and Amit Sheth. 2018.
\newblock A quality type-aware annotated corpus and lexicon for harassment
  research.
\newblock In \emph{Proceedings of the 10th ACM Conference on Web Science},
  pages 33--36.

\bibitem[{Ribeiro et~al.(2018)Ribeiro, Calais, Santos, Almeida, and
  Meira~Jr}]{ribeiro2018characterizing}
Manoel Ribeiro, Pedro Calais, Yuri Santos, Virg{\'\i}lio Almeida, and Wagner
  Meira~Jr. 2018.
\newblock Characterizing and detecting hateful users on twitter.
\newblock In \emph{Proceedings of the International AAAI Conference on Web and
  Social Media}, volume~12.

\bibitem[{Saha et~al.(2019)Saha, Chandrasekharan, and
  De~Choudhury}]{saha2019prevalence}
Koustuv Saha, Eshwar Chandrasekharan, and Munmun De~Choudhury. 2019.
\newblock Prevalence and psychological effects of hateful speech in online
  college communities.
\newblock In \emph{Proceedings of the 10th ACM Conference on Web Science},
  pages 255--264.

\bibitem[{Waseem(2016)}]{waseem2016you}
Zeerak Waseem. 2016.
\newblock Are you a racist or am i seeing things? annotator influence on hate
  speech detection on twitter.
\newblock In \emph{Proceedings of the first workshop on NLP and computational
  social science}, pages 138--142.

\bibitem[{Ziems et~al.(2020)Ziems, He, Soni, and Kumar}]{ziems2020racism}
Caleb Ziems, Bing He, Sandeep Soni, and Srijan Kumar. 2020.
\newblock Racism is a virus: Anti-asian hate and counterhate in social media
  during the covid-19 crisis.
\newblock \emph{arXiv preprint arXiv:2005.12423}.

\end{thebibliography}

\appendix

\end{document}